\documentclass{article}

\usepackage[T1]{fontenc}
\usepackage{pearl_preprint,times}
\usepackage{amsmath,amsfonts,bm}

\def\eqref#1{equation~\ref{#1}}
\def\1{\bm{1}}

\DeclareMathAlphabet{\mathsfit}{\encodingdefault}{\sfdefault}{m}{sl}
\SetMathAlphabet{\mathsfit}{bold}{\encodingdefault}{\sfdefault}{bx}{n}

\usepackage{amsmath,amssymb,mathtools}
\usepackage{booktabs}
\usepackage{multirow}
\usepackage{array}
\usepackage{graphicx}
\usepackage{float}
\usepackage{wrapfig}
\usepackage{microtype}
\usepackage{xcolor}
\usepackage{colortbl}
\usepackage{url}
\usepackage{enumitem}
\usepackage{placeins}
\usepackage{needspace}
\usepackage{upquote}
\usepackage{listings}
\usepackage{tcolorbox}
\tcbuselibrary{breakable,listings}
\usepackage[hidelinks]{hyperref}

\newcommand{\method}{PEARL}
\newcommand{\arpo}{ARPO}

\title{PEARL: Process-Experience Aware Reinforcement Learning for Multi-Turn Agents}

\author{
\makebox[\dimexpr\textwidth-2\tabcolsep-3pt\relax][c]{%
\begin{tabular}{c}
Jie Yang\textsuperscript{1,}\thanks{Equal contribution.},\quad
Mianqiu Huang\textsuperscript{1,}\footnotemark[1],\quad
Yiwei Zhao\textsuperscript{1,}\footnotemark[1]\\[2pt]
Bo Wang\textsuperscript{1},\quad
Yang Liu\textsuperscript{1},\quad
Yuxin Wang\textsuperscript{2,}\thanks{Corresponding authors: Yuxin Wang (wangyuxin@sii.edu.cn)}\\[5pt]
\normalfont\textsuperscript{1}Fudan University,\quad
\textsuperscript{2}Shanghai Innovation Institute
\end{tabular}}
}

\hypersetup{
  pdfauthor={Jie Yang, Mianqiu Huang, Yiwei Zhao, Bo Wang, Yang Liu, Yuxin Wang},
  pdftitle={PEARL: Process-Experience Aware Reinforcement Learning for Multi-Turn Agents}
}

\pearlpreprintcopy

\begin{document}

\maketitle
\lhead{}

\begin{abstract}
Terminal rewards make multi-turn agents trainable without step-level annotations, but group-relative advantages provide little guidance about which intermediate decisions to adjust.
Failed trajectories also contain reasonable actions, while unfavorable decision patterns can recur as training proceeds.
We introduce \textbf{\method{}} (Process-Experience Aware Reinforcement Learning), which improves local credit assignment without an additional critic.
Given two continuations from the same history with different outcomes, \method{} locates their first divergence and builds a feature representation of each branch's local decision.
It clusters these decision features in recent failure and success memories to identify recurring local patterns.
Using this pattern memory, \method{} selectively strengthens negative advantages at lower-reward decisions, while leaving positive advantages and the base policy optimizer unchanged.
On AppWorld, \method{} improves over \arpo{} on Normal and Challenge tasks at both 4B and 8B scales.
On ALFWorld, it achieves 97.9\% mean success at 4B, 2.7 points above the strongest baseline, while remaining competitive at 8B.
Ablations of shaping location, success-bank filtering, and memory window support the value of selective, recurring-pattern-based updates.
\end{abstract}

\section{Introduction}
\label{sec:introduction}

Multi-turn language agents can be trained from verifiable task outcomes without annotating every intermediate action.
The resulting reward, however, says little about which decisions along a long interaction deserve adjustment.
Group Relative Policy Optimization (GRPO) replaces a learned critic with group-relative outcome advantages, but tokens in the same trajectory share the same reward-based comparison~\citep{shao2024deepseekmath}.
Actor--critic methods such as PPO can estimate state-dependent values, but require training an additional critic and do not by themselves provide labels for locally correct or incorrect actions~\citep{schulman2017proximal}.
The first challenge is therefore to use step-level information from successful and failed trajectories to improve credit assignment without adding a critic or changing the underlying policy optimizer.

The second challenge is to decide which local experiences are worth reusing.
Similar decisions may recur in lower-reward trajectories, but a failed trajectory also contains reasonable actions that should not receive extra penalties.
Moreover, the policy changes during training, so patterns observed early may become less relevant to its current behavior.
Repeated failure-side occurrence alone is thus insufficient: useful supervision must identify recurring outcome-associated decisions while accounting for successful behavior and the changing policy.

Branching rollouts provide a starting point for both challenges by generating alternative continuations from a shared interaction history~\citep{dong2026arpo,ji2026treesearch}.
When their outcomes differ, the branches expose a local decision contrast without requiring a learned value estimate.
Yet one contrast does not prove that the differing action caused the outcome; the question is how to reuse such contrasts across trajectories without penalizing behaviors also seen in successful continuations.

\begin{figure}[t]
\centering
\includegraphics[width=\textwidth]{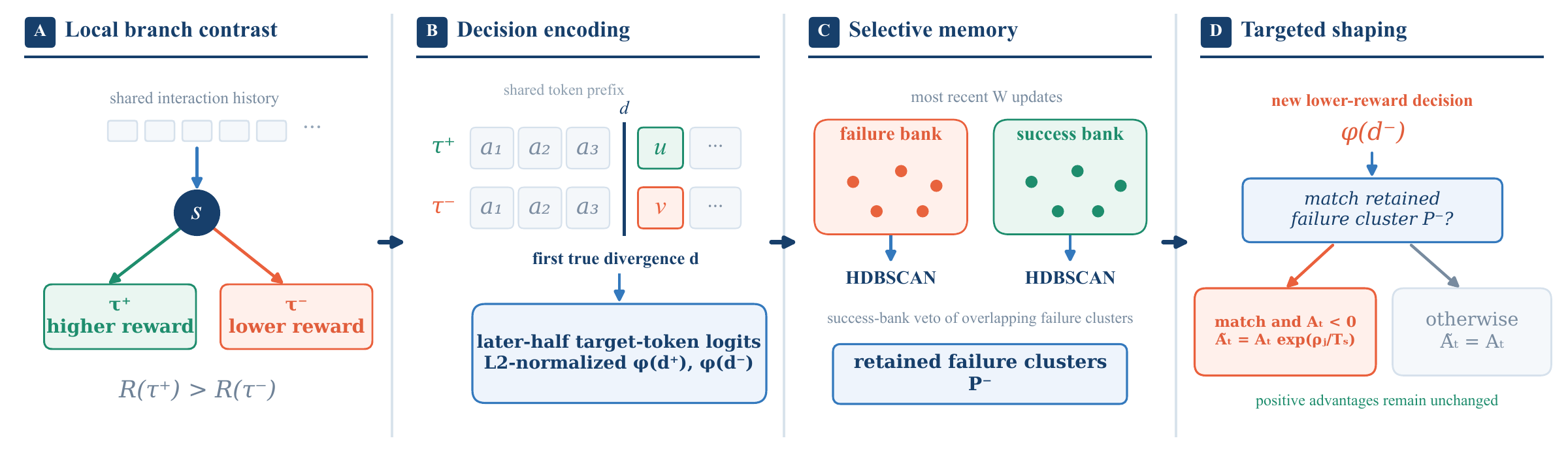}
\caption{Overview of \method{}. Unequal-outcome branches provide local decision contrasts. Recurring decision patterns are stored in recent failure and success memories, success-side evidence filters ambiguous failure patterns, and retained failure patterns guide stronger negative updates.}
\label{fig:pipeline}
\end{figure}
\FloatBarrier

We introduce \textbf{\method{}} (Process-Experience Aware Reinforcement Learning), illustrated in Figure~\ref{fig:pipeline}.
For branches with different outcomes, \method{} localizes their first actual divergence and represents each selected decision with layer-wise token logits.
It clusters these features across updates in recent failure and success banks, so recurring patterns can inform later training decisions rather than being used only once.
Success-side matches exclude ambiguous failure clusters, while a sliding window limits the influence of older experience.
Only negative advantages at decisions matching retained failure-side patterns are strengthened; positive advantages and the base policy optimizer remain unchanged.

The retained patterns are candidates for targeted updates, not labels of intrinsically incorrect actions.
Their value comes from combining a local outcome contrast with recurrence and a check against successful behavior.

We evaluate \method{} on AppWorld~\citep{trivedi2024appworld} and ALFWorld~\citep{shridharalfworld} at both 4B and 8B model scales.
On AppWorld, \method{} improves over \arpo{} on both Normal and Challenge tasks at both scales.
On ALFWorld, \method{} achieves the best reported mean at 4B and remains competitive at 8B.
Controlled ablations compare pattern-guided with random shaping and measure the effect of success-bank filtering, while process diagnostics track how often the resulting memory activates shaping.

Our contributions are:
\begin{itemize}[leftmargin=*,itemsep=1pt,topsep=3pt]
    \item We turn terminal-reward differences between continuations from a shared history into local training signals, improving credit assignment without step-level annotations or an additional critic.
    \item We develop a selective memory for recurring decision patterns that uses successful experience and recent observations to focus extra negative updates, rather than applying them to every action in failed trajectories.
    \item We show that \method{} improves over \arpo{} on AppWorld and ALFWorld at both 4B and 8B scales; controlled ablations further examine the roles of targeted shaping and success-bank filtering.
\end{itemize}

\section{Related Work}
\label{sec:related_work}

\paragraph{Group-relative and branching agent RL.}
GRPO replaces a learned critic with within-group reward normalization and has become a common foundation for reasoning RL.
Agent-oriented extensions exploit additional rollout structure: GiGPO combines within- and across-group comparisons~\citep{fenggroup}; Tree-GRPO estimates intra-tree and inter-tree advantages from shared-prefix rollouts; and ARPO allocates branches using uncertainty changes along an interaction.
Sequence-level normalization and strategic trajectory abstractions provide related ways to restructure the optimization signal~\citep{zheng2025group,xue2026strata}.
These methods determine which trajectories to sample or how their current rewards become advantages.
In contrast, \method{} uses the resulting branches to build a cross-update memory of recurrent decisions and remains an add-on to the underlying branching optimizer.

\paragraph{Credit assignment from sparse outcome rewards.}
Long-horizon agent training must infer useful local updates from trajectory-level outcomes.
Recent approaches improve this inference through future-sensitive objectives, selective eligibility traces, information gain, entropy balancing, or learned process abstractions~\citep{ma2026fipo,mou2026beyond,wang2025information,dong2025agentic}.
Hindsight Policy Optimization (HPO) derives step-level advantages without a critic by comparing current and success-conditioned hindsight distributions in a semantic intent space via Wasserstein distance~\citep{jiang2026outcomes}.
These methods construct finer action-level signals; \method{} addresses the complementary question of which locally contrasted decisions recur on lower-reward branches without corresponding success-side support.
It compares continuations from a shared branch point, aggregates matching decisions across updates, and strengthens negative advantages only after recurrence and success-disambiguation tests.

\paragraph{Memory-based reward shaping and internal representations.}
The closest representation-level precursor is MEDS, which clusters layer-wise target-token logits and uses their recurrence for dynamic reward shaping~\citep{liu2026past}.
\method{} retains the useful idea that a token's logit trajectory across layers is a compact signature of a decision, but changes both supervision and memory semantics.
Specifically, outcome-side labels come from discordant parent--branch pairs rather than whole-trajectory outcomes; regions start at the first true divergence; failure and success representations occupy separate banks; and success acts as an asymmetric filter before negative-only shaping. These labels identify candidate update locations, not ground-truth local errors.
HDBSCAN~\citep{mcinnes2017hdbscan} permits an unknown number of patterns and leaves low-density observations as noise, while a sliding window limits the lifetime of representations collected under older policies.
Thus, \method{} is a training-time memory for selective credit assignment rather than an inference-time store of demonstrations or reflections.

\section{Method}
\label{sec:method}

\subsection{Overview}

For a prompt $x$, an agent policy $\pi_\theta$ generates a multi-turn trajectory
$\tau=(a_1,o_1,\ldots,a_T,o_T)$ and receives a verifiable terminal reward $R(\tau)$.
Our base optimizer is ARPO, which samples multiple trajectories for the same prompt and may branch from an intermediate state to explore alternative continuations.

\method{} augments ARPO with a cross-update memory that turns recurring lower-reward-side decision patterns into additional training signals.
As illustrated in Figure~\ref{fig:pipeline}, the method has four stages.
First, branches with different outcomes provide local failure--success comparisons.
Second, each divergent decision is represented by a compact decision-aligned feature.
Third, recent failure and success features are clustered separately, and success-side evidence is used to filter ambiguous failure patterns.
Finally, negative advantages are strengthened only for decisions matching the retained failure patterns.

\subsection{Local supervision from discordant branches}
\label{sec:counterfactual}

A terminal failure reward does not reveal which local decisions should receive stronger negative updates.
Branches from the same saved state provide a more focused comparison because they share the same preceding interaction history.

We therefore use only parent--branch pairs whose terminal rewards differ.
For a parent trajectory $\tau^p$ and its branch $\tau^b$, we define the higher-reward continuation as successful and the lower-reward continuation as failed:
\begin{equation}
  (\tau^+,\tau^-)=
  \begin{cases}
  (\tau^b,\tau^p), & R(\tau^b)>R(\tau^p),\\
  (\tau^p,\tau^b), & R(\tau^p)>R(\tau^b).
  \end{cases}
  \label{eq:pair-label}
\end{equation}
Pairs with equal outcomes contribute no memory entries.

Because the two continuations may still share tokens after branching, we locate their first actual token difference $d$ and use only the remainder of that action as the local comparison region.
Pairs without a valid divergence are skipped.
This removes the shared prefix from supervision and focuses the comparison on where the two continuations begin to differ.

\subsection{Decision-aligned layer-logit representation}
\label{sec:representation}

We next encode the decision at the first divergence.
Instead of clustering full hidden states, which contain substantial contextual information unrelated to the selected action, we represent how strongly different layers support the token that is actually chosen.

\begin{wrapfigure}{r}{0.45\textwidth}
\vspace{-8pt}
\centering
\includegraphics[width=0.44\textwidth]{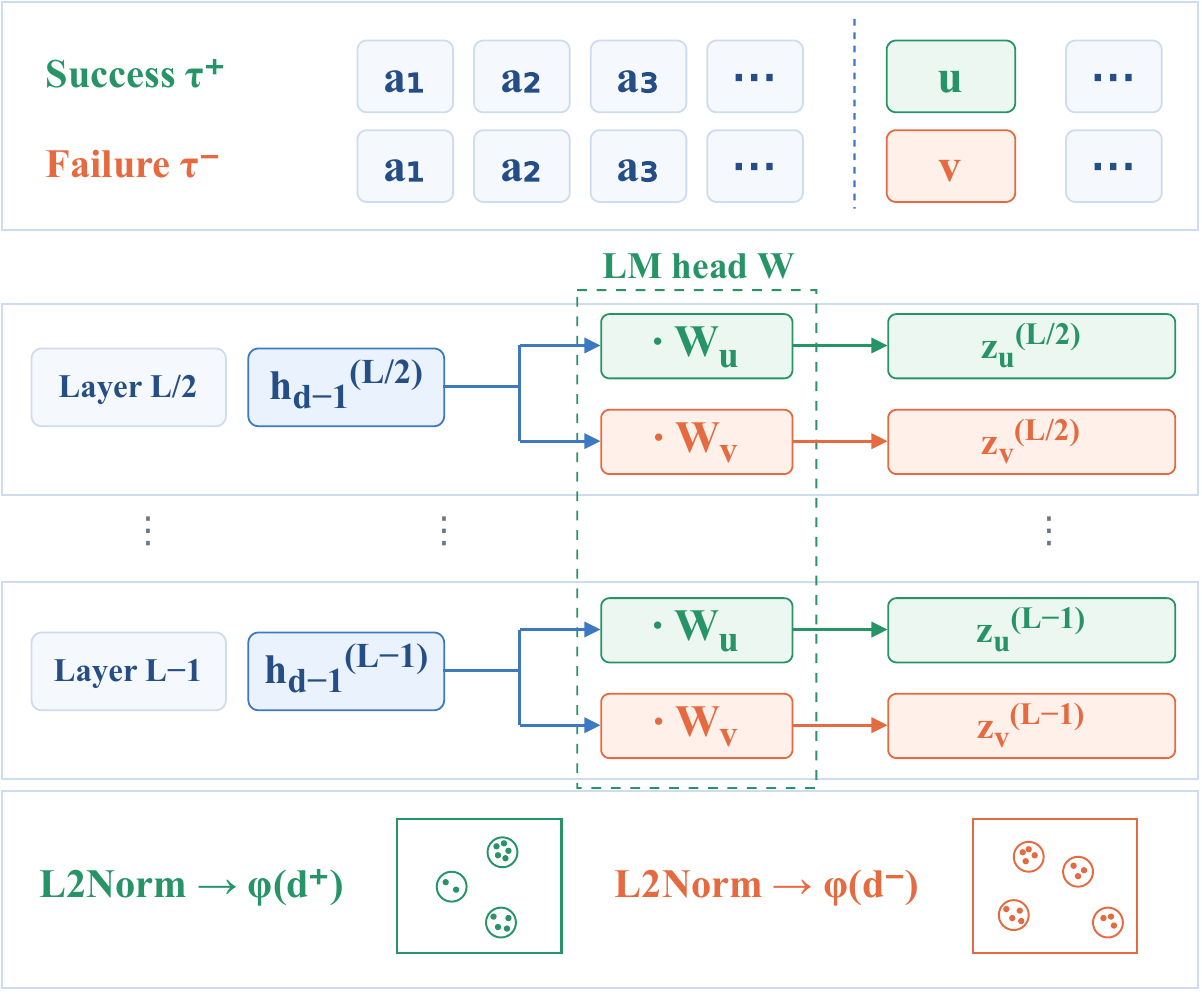}
\caption{Decision representation at the first divergence. Each branch scores its selected token across the latter half of the network, producing a compact layer-logit feature for clustering.}
\label{fig:decision-feature-detail}
\vspace{-8pt}
\end{wrapfigure}

Let $y_d$ denote the token at divergent position $d$, $h_{d-1}^{(\ell)}$ the hidden state used to predict it at layer $\ell$, and $W_{y_d}$ the corresponding output-head row.
We compute
\begin{equation}
  z_d^{(\ell)}=
  \left\langle h_{d-1}^{(\ell)},W_{y_d}\right\rangle,
\end{equation}
and concatenate the target-token logits from the latter half of an $L$-layer model:
\begin{equation}
  \phi(d)=\operatorname{L2Norm}
  \left([z_d^{(L/2)},\ldots,z_d^{(L-1)}]\right).
  \label{eq:meds}
\end{equation}

For Qwen3-4B-Instruct-2507, $L=36$, yielding an 18-dimensional representation compared with the 2,560-dimensional final hidden state.
The resulting feature captures how support for the selected token develops across later layers while remaining compact for density-based clustering.
Section~\ref{sec:representation-ablation} compares this representation with several alternatives.

\subsection{Clustering and filtering recent decision patterns}
\label{sec:banks}

We maintain separate failure and success banks to identify recurring decision patterns.
At update $s$, each discordant pair contributes a feature $\phi(d^-)$ from the lower-reward branch and $\phi(d^+)$ from the higher-reward branch:
\[
\phi(d^-)\rightarrow\mathcal{B}_s^-,
\qquad
\phi(d^+)\rightarrow\mathcal{B}_s^+.
\]

Because the policy changes during training, both banks retain only recent observations.
Entries older than $W$ updates are removed.
We periodically fit HDBSCAN to the L2-normalized features in each bank, allowing recurrent patterns to form clusters while isolated observations remain noise.
We use $W=25$ and refit every $K=10$ updates by default.
The default clustering configuration is MCS10/MS5.

Recurring failure patterns are not automatically suitable for shaping, since the same behavior may also appear in successful trajectories.
We therefore use the success bank to filter the failure clusters.
Let $\mathcal{C}_s^-=\{C_j^-\}$ denote the failure clusters and $\mu_j^-$ the centroid of cluster $j$.
A failure cluster is retained only if its centroid is not assigned to a success cluster:
\begin{equation}
  \mathcal{P}_s^-=
  \left\{
  C_j^-\in\mathcal{C}_s^-:
  \operatorname{predict}_{\mathcal{B}_s^+}(\mu_j^-)=-1
  \right\}.
  \label{eq:hardfilter}
\end{equation}

This filtering is asymmetric: successful behavior is used to reject ambiguous failure patterns, while failure patterns do not remove success clusters.

\subsection{Selective negative-advantage shaping}
\label{sec:shaping}

The retained failure clusters provide additional supervision for recurring lower-reward decision patterns.
We use them only to strengthen negative updates, while leaving positive advantages unchanged.

For a current lower-reward decision $\phi(d^-)$ that matches retained failure cluster
$C_j^-\in\mathcal{P}_s^-$, we define its relative cluster support as
\begin{equation}
\rho_j=
\frac{|C_j^-|}
{\sum_{C_k^-\in\mathcal{P}_s^-}|C_k^-|}.
\end{equation}

Let $A_t$ denote the original token advantage and $T_s$ the shaping temperature, which is set to $1$ by default.
For tokens from the localized divergence $d^-$ to the end of the corresponding action, we use
\begin{equation}
\widetilde{A}_t=
\begin{cases}
A_t\exp\!\left(\rho_j/T_s\right),
& A_t<0\ \text{and }\phi(d^-)\in C_j^-,\\
A_t, & \text{otherwise}.
\end{cases}
\label{eq:shaping}
\end{equation}
Tokens before the divergence and all positive advantages remain unchanged.

The base clipped policy-gradient objective is then applied using $\widetilde{A}_t$:
\begin{equation}
  \mathcal{L}_{\text{PG}}
  =
  -\mathbb{E}_t\!\left[
  \min\left(
  r_t\widetilde{A}_t,
  \operatorname{clip}(r_t,1-\epsilon,1+\epsilon)\widetilde{A}_t
  \right)
  \right],
\end{equation}
where
$r_t=\pi_\theta(y_t\mid s_t)/\pi_{\theta_{\mathrm{old}}}(y_t\mid s_t)$.

Thus, \method{} leaves the underlying policy optimization procedure unchanged and modifies only the magnitude of selected negative updates.

\section{Experiments}
\label{sec:experiments}

We evaluate \method{} on AppWorld and ALFWorld, followed by controlled ablations and process diagnostics.

\subsection{Experimental setup}

\paragraph{Benchmarks and models.}
We evaluate Qwen3-4B-Instruct-2507 and Qwen3-8B~\citep{yang2025qwen3} on two multi-turn agent benchmarks.
AppWorld evaluates interactive coding agents on Normal and more compositional Challenge tasks over simulated applications, while ALFWorld evaluates text agents on Seen and Unseen embodied household tasks.
Success rate is the primary metric for both benchmarks.
The agent prompts for both environments are provided in Appendix~\ref{app:prompts}.

\paragraph{Baselines.}
We compare with GRPO, GiGPO, Tree-GRPO, and ARPO under the same model and environment infrastructure.
\method{} is built on ARPO because its branching rollouts provide the matched local comparisons required by Section~\ref{sec:counterfactual}.
GRPO, GiGPO, ARPO, and \method{} use 16 rollouts per prompt; Tree-GRPO follows its tree-expansion setting and produces 24 physical leaves per prompt group.

\paragraph{Training and evaluation protocol.}
We evaluate AppWorld after 400 optimizer updates and ALFWorld after 300 updates.
All main experiments use three training seeds, and we report the mean and standard deviation across seeds.
The default \method{} configuration uses MCS10/MS5, $W=25$, $K=10$, $T_s=1$, hard success-bank filtering, and failure-only advantage shaping.
Unless otherwise specified, ablations are conducted on AppWorld with Qwen3-4B-Instruct-2507 using a single controlled run and evaluated at step 400.

\subsection{Main results}
\label{sec:main-results}

\begin{table*}[t]
\centering
\caption{Final task success rates (\%) after 400 AppWorld or 300 ALFWorld updates. Each ``Mean'' equally weights the two splits. Trained-method values are mean $\pm$ standard deviation over three runs; Base Model is the untrained model. Bold and underline mark the best and second-best distinct values within each model scale, with ties formatted alike.}
\label{tab:main-results}
\setlength{\tabcolsep}{4.5pt}
\renewcommand{\arraystretch}{1.28}
\begin{tabular}{@{}lrrrrrr@{}}
\toprule
& \multicolumn{3}{c}{AppWorld} & \multicolumn{3}{c}{ALFWorld} \\
\cmidrule(lr){2-4}\cmidrule(lr){5-7}
Method & Normal & Challenge & Mean & Seen & Unseen & Mean \\
\midrule
\multicolumn{7}{@{}c@{}}{Qwen3-4B-Instruct-2507} \\
Base Model & 8.93 & 5.60 & 7.26 & 38.1 & 33.1 & 35.6 \\
GRPO      & 52.38$\pm$0.79 & 24.38$\pm$0.16 & 38.38$\pm$0.48 & 94.9$\pm$2.8 & 94.3$\pm$5.7 & 94.6$\pm$4.2 \\
ARPO      & \underline{53.37}$\pm$2.18 & \underline{29.02}$\pm$2.56 & \underline{41.19}$\pm$2.37 & 95.1$\pm$0.4 & 92.3$\pm$0.8 & 93.7$\pm$0.2 \\
GiGPO     & 44.84$\pm$4.76 & 24.02$\pm$0.52 & 34.43$\pm$2.64 & \underline{96.0}$\pm$2.0 & \underline{94.4}$\pm$3.4 & \underline{95.2}$\pm$2.7 \\
Tree-GRPO & 36.41$\pm$1.49 & 18.98$\pm$0.04 & 27.70$\pm$0.72 & 85.6$\pm$10.1 & 79.9$\pm$14.2 & 82.7$\pm$12.1 \\
\rowcolor[gray]{0.94}
\textbf{\method{}} & \textbf{57.04}$\pm$0.69 & \textbf{30.18}$\pm$0.84 & \textbf{43.61}$\pm$0.77 & \textbf{98.0}$\pm$0.6 & \textbf{97.9}$\pm$0.4 & \textbf{97.9}$\pm$0.5 \\
\midrule
\multicolumn{7}{@{}c@{}}{Qwen3-8B} \\
Base Model & 11.31 & 4.48 & 7.89 & 57.6 & 55.5 & 56.5 \\
GRPO      & 43.95$\pm$0.10 & 20.26$\pm$2.84 & 32.11$\pm$1.47 & 94.9$\pm$1.8 & \underline{96.4}$\pm$1.1 & 95.6$\pm$1.5 \\
ARPO      & \underline{50.60}$\pm$1.98 & \underline{23.70}$\pm$3.96 & \underline{37.15}$\pm$2.97 & \underline{95.5}$\pm$0.3 & 95.4$\pm$0.6 & 95.4$\pm$0.2 \\
GiGPO     & 45.24$\pm$4.37 & 17.91$\pm$0.08 & 31.57$\pm$2.14 & \textbf{97.7}$\pm$0.9 & \textbf{98.0}$\pm$0.0 & \textbf{97.9}$\pm$0.4 \\
Tree-GRPO & 42.16$\pm$2.48 & 16.67$\pm$0.76 & 29.41$\pm$1.62 & 95.0$\pm$0.2 & 93.8$\pm$0.3 & 94.4$\pm$0.2 \\
\rowcolor[gray]{0.94}
\textbf{\method{}} & \textbf{55.16}$\pm$1.98 & \textbf{25.10}$\pm$1.76 & \textbf{40.13}$\pm$1.87 & \textbf{97.7}$\pm$1.2 & \underline{96.4}$\pm$1.9 & \underline{97.1}$\pm$1.6 \\
\bottomrule
\end{tabular}

\end{table*}

Table~\ref{tab:main-results} summarizes the main results.
Across both benchmarks and model scales, \method{} improves mean success over its ARPO backbone.
On AppWorld, it leads both splits at both scales, with Normal/Challenge gains over ARPO of 3.67/1.16 points at 4B and 4.56/1.40 points at 8B.

\begin{figure*}[t]
\centering
\includegraphics[width=0.98\textwidth]{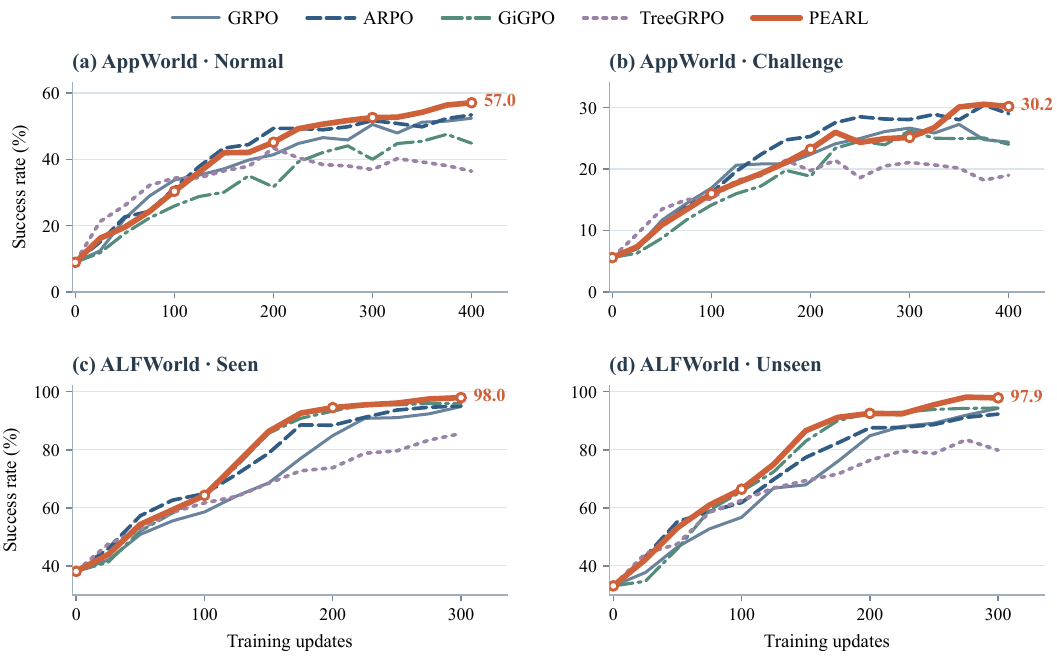}
\caption{Qwen3-4B-Instruct-2507 evaluation success rates (\%) on AppWorld (top, through 400 updates) and ALFWorld (bottom, through 300 updates). Step 0 is the shared pre-training evaluation; subsequent points, spaced every 25 updates, are averages across the available training runs. Orange endpoint labels show PEARL's final scores reported in Table~\ref{tab:main-results}.}
\label{fig:eval-trajectories}
\end{figure*}

On ALFWorld, \method{} achieves the highest 4B mean at 97.9\%, 2.7 points above GiGPO, the strongest baseline.
At 8B, its 97.1\% mean exceeds ARPO by 1.7 points but trails GiGPO by 0.8 points.
Figure~\ref{fig:eval-trajectories} shows the corresponding 4B evaluation trajectories throughout training.

\subsection{Ablations}
\label{sec:ablation}

We next examine the main design choices of \method{}, including shaping direction, shaping location, success-bank filtering, memory window, and clustering features.

\begin{figure}[t]
\centering
{
\textcolor[HTML]{ACCEE3}{\rule{9pt}{6pt}}\ Normal\qquad
\textcolor[HTML]{377DA8}{\rule{9pt}{6pt}}\ Challenge\par}
\smallskip
\begin{minipage}[t]{0.32\linewidth}
\centering
\includegraphics[width=\linewidth]{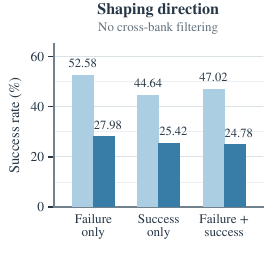}
\par (a)
\end{minipage}\hfill
\begin{minipage}[t]{0.28\linewidth}
\centering
\includegraphics[width=\linewidth]{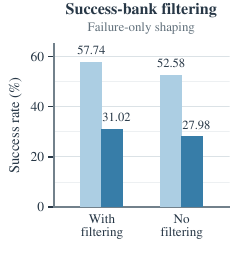}
\par (b)
\end{minipage}\hfill
\begin{minipage}[t]{0.36\linewidth}
\centering
\includegraphics[width=\linewidth]{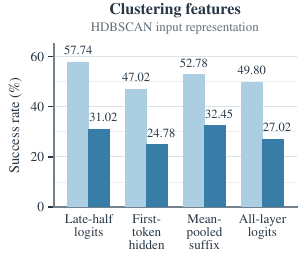}
\par (c)
\end{minipage}
\caption{Single-run AppWorld success rates (\%) at step 400 with Qwen3-4B-Instruct-2507. (a) Failure-only, success-only, and bidirectional shaping without cross-bank filtering. (b) With versus without success-bank filtering, holding failure-only shaping and MCS10/MS5 fixed. (c) Four HDBSCAN input representations, holding all other settings fixed. Light and dark bars denote Normal and Challenge, respectively.}
\label{fig:ablation-components}
\end{figure}

\paragraph{Effect of shaping direction.}
Negative and positive shaping are not interchangeable.
Figure~\ref{fig:ablation-components}(a) varies the shaping direction without cross-bank filtering.
Failure-only shaping achieves 52.58\%/27.98\% on Normal/Challenge, compared with 44.64\%/25.42\% for success-only shaping and 47.02\%/24.78\% for bidirectional shaping.
Adding success shaping to failure shaping lowers performance by 5.56/3.20 points.
Thus, among these shaping directions, strengthening negative updates on recurrent lower-reward-side patterns is more effective than adding positive weight to success-side patterns.
Section~\ref{sec:process-diagnostics} examines the corresponding shaping activity and response length.

\paragraph{Success evidence helps disambiguate failure patterns.}
Patterns observed on failed branches may also occur in successful behavior and are therefore poor targets for additional negative shaping.
To isolate the success bank's filtering role, we compare failure-only shaping with and without the cross-bank veto while keeping both banks and the clustering settings unchanged.
As shown in Figure~\ref{fig:ablation-components}(b), enabling this filter improves Normal/Challenge performance by 5.16/3.04 points over no filtering.
This result supports using successful experience to reject ambiguous failure patterns before applying extra negative weight.
Section~\ref{sec:memory-case-study} examines a concrete example of this filtering.

\paragraph{Pattern-guided shaping outperforms matched random locations.}
We test whether the location of extra negative weight matters, rather than just its presence.
The control retains the same discordant-branch labels, banks, clustering, and success-bank filter as \method{}: a retained failure-cluster match still triggers shaping and determines its multiplier.
For each match, \method{} amplifies the negative-advantage tokens between the first divergent token and the end of that action.
The random-location control counts these tokens, then uniformly samples the same number of negative-advantage action tokens from anywhere in the same lower-reward trajectory and applies the same multiplier.
For example, if a matched action contains four eligible tokens, both variants amplify four tokens with that match's multiplier; only the control chooses their positions at random.
Thus, the comparison changes where each matched shaping event is applied, not its token budget or weight.
At step 400, pattern-guided shaping achieves 57.74\%/31.02\% on Normal/Challenge, compared with 49.17\%/25.37\% for random shaping.
With each match's token budget and multiplier controlled, the 8.57/5.65-point gains show that PEARL identifies more effective locations for negative updates than random selection within the same trajectory.

\paragraph{Recent memory outperforms unlimited history.}
The bank must supply patterns relevant to the current policy, so we test whether retaining only recent entries is more effective than keeping the entire training history.
With hard filtering and otherwise identical settings, extending the window from 25 updates to unlimited history ($W=\infty$) reduces Normal performance from 57.74\% to 50.00\% and Challenge performance from 31.02\% to 26.38\%.
Thus, retaining more historical experience does not necessarily provide a better shaping signal.
Section~\ref{sec:process-diagnostics} examines how the memory window affects shaping activity during training.

\paragraph{Ablation on clustering features.}
\label{sec:representation-ablation}
We test which feature representation is most effective for modeling local decisions in PEARL.
We compare four HDBSCAN inputs while keeping all other settings fixed: late-half target logits (18 dimensions), the final normalized hidden state after the first divergent token has been processed (2,560 dimensions), mean-pooled final hidden states over the divergent action suffix (2,560 dimensions), and all-layer target logits (36 dimensions).
The first-token hidden-state baseline thus uses the post-token state, not the shared pre-token state used to predict that token.
All variants use MCS10/MS5 and Euclidean distance on L2-normalized features.

As shown in Figure~\ref{fig:ablation-components}(c), late-half target logits achieve 57.74\% on Normal and 31.02\% on Challenge.
They give the best Normal result while remaining within 1.43 points of the best Challenge result.
Compared with first-token hidden states, late-half logits improve Normal/Challenge by 10.72/6.24 points, and compared with all-layer logits by 7.94/4.00 points.
Mean-pooled hidden states obtain the highest Challenge score, but their Normal performance drops to 52.78\%.
Among these complete representation choices, late-half target logits provide the strongest balance across the two splits at step 400 and are therefore used as the default clustering feature.

\subsection{Process diagnostics: how does PEARL use its memory?}
\label{sec:process-diagnostics}

The ablations above establish how PEARL's design choices affect final performance.
We next examine how memory retention and filtering translate into shaping activity, and why the resulting update is negative-only.

\begin{figure}[t]
\centering
\includegraphics[width=\linewidth]{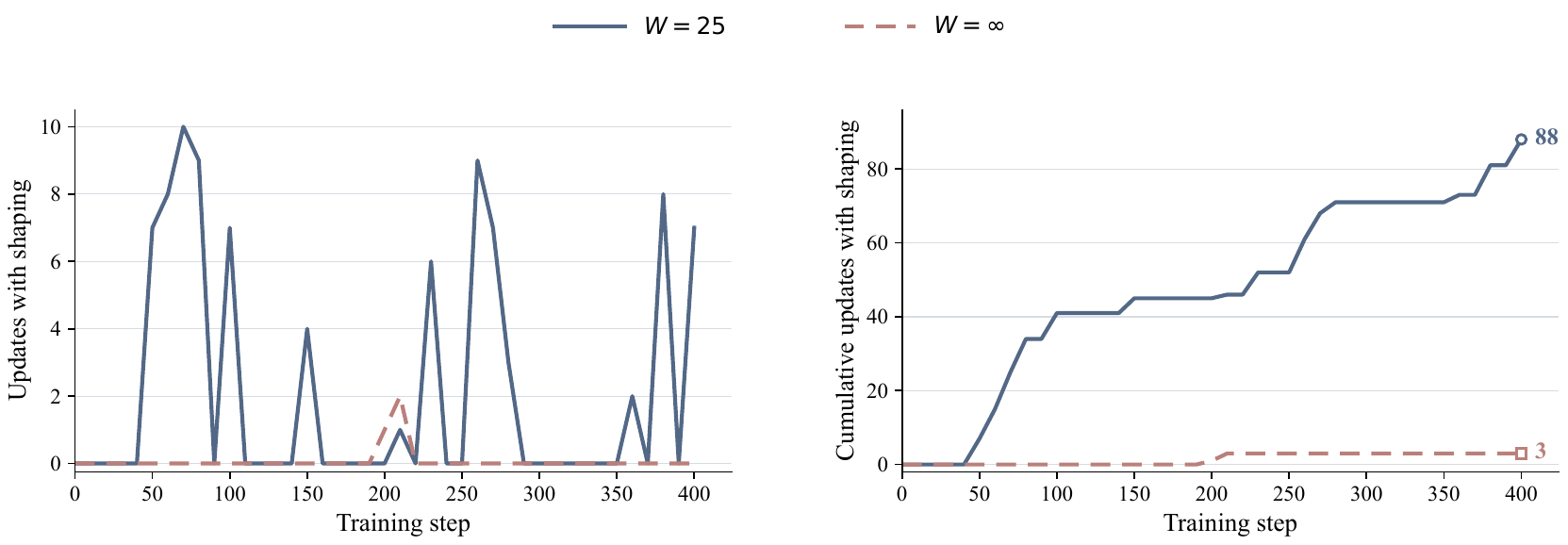}
\caption{Failure-shaping activity with a 25-update window and unlimited history ($W=\infty$) on AppWorld (Qwen3-4B-Instruct-2507, hard filtering, MCS10/MS5; one run per setting). Left: number of updates with active failure shaping in each consecutive block of 10 updates. Right: cumulative number of updates with active failure shaping, reaching 88 for $W=25$ and 3 for unlimited history.}
\label{fig:window-shaping}
\end{figure}

\paragraph{Finite history preserves more usable failure signals.}
Figure~\ref{fig:window-shaping} shows that $W=25$ activates failure shaping in 88 of 400 updates, compared with only 3 under unlimited history.
Unlimited history keeps features from earlier policies in both banks.
At update 400, one retained failure cluster contains 1,200 of 1,368 failure entries, consistent with accumulated entries obscuring smaller recent patterns in the density structure; at updates 100 and 300, success-bank filtering leaves no retained failure cluster (Appendix Table~\ref{tab:w0-process}).
Even a large retained cluster shapes an update only when a current decision matches it.
These snapshots show why bank size alone is misleading: a dominant cluster or a cross-bank veto can leave few patterns that match current decisions.

\paragraph{Positive shaping reinforces an already rewarded signal.}
The base objective already favors higher-reward continuations; success shaping adds weight to recurring decisions within them, whereas failure shaping strengthens penalties on recurring lower-reward decisions.
In the no-filter ablation, the positive signal is far from inactive: success-only shaping activates on 65\% of the final 100 updates, compared with 40\% for failure-only shaping, yet achieves lower task success.
When success shaping is added to failure shaping, mean evaluation response length rises from 4,145 to 7,770 tokens on Normal and from 4,367 to 8,505 on Challenge, while success declines.
The extra positive weight therefore changes generated behavior in the combined run without improving task completion, motivating PEARL's focus on recurrent lower-reward decisions.

\subsubsection{Case study: success evidence vetoes shared behavior}
\label{sec:memory-case-study}
A failed trajectory can contain useful shared actions. Tables~\ref{tab:appworld-paired-case} and~\ref{tab:alfworld-paired-case} show how success memory vetoes such actions in AppWorld and ALFWorld, respectively.

\begin{table}[H]
\centering
\caption{AppWorld, Qwen3-4B-Instruct-2507, training step 400. Two behaviors from each cluster exhibit the same API-inspection pattern; success cluster $S_1$ filters failure cluster $F_1$.}
\label{tab:appworld-paired-case}
\setlength{\tabcolsep}{4pt}
\renewcommand{\arraystretch}{1.02}
\begin{tabular}{@{}>{\raggedright\arraybackslash}p{0.14\linewidth}>{\raggedright\arraybackslash}p{\dimexpr0.86\linewidth-8pt\relax}@{}}
\toprule
Cluster & Trajectory behavior and outcome \\
\midrule
\textbf{Success}\par $S_1$ & \begin{tabular}[t]{@{}>{\raggedright\arraybackslash}p{\linewidth}@{}}
\textbf{Behavior 1: friends and roommates.} Inspect the contact-relationship API specification; receive a valid schema. Later, collect the matching requests before denying them, and reject all 14. The trajectory succeeds. \\[2pt]
\midrule
\textbf{Behavior 2: siblings.} Inspect the same API specification; receive a valid schema. Later, check both transaction senders and receivers and like all 8 matching transactions. The trajectory succeeds. \\
\end{tabular} \\[3pt]
\midrule
\textbf{Failure}\par $F_1$ & \begin{tabular}[t]{@{}>{\raggedright\arraybackslash}p{\linewidth}@{}}
\textbf{Behavior 1: friends and roommates.} Inspect the same specification; receive a valid schema.\par\textbf{Later failure:} Deny requests while paging through the pending list; reject only 13 of 14, missing request 1844. Modifying the list during traversal is consistent with a pagination skip. \\[2pt]
\midrule
\textbf{Behavior 2: siblings.} Inspect the same specification; receive a valid schema.\par\textbf{Later failure:} Check only whether the sender is a sibling, ignoring sibling receivers. Like only 2 of the 8 matching transactions. \\
\end{tabular} \\[3pt]
\midrule
\multicolumn{2}{@{}p{\linewidth}@{}}{\textbf{Shared pattern.} Inspect the contact-relationship API specification before identifying people by relationship.} \\[3pt]
\multicolumn{2}{@{}p{\linewidth}@{}}{\textbf{Memory decision.} \textbf{$S_1$ filters $F_1$.}  The same specification-inspection behavior occurs in successful trajectories. The corresponding failure cluster is excluded from additional negative shaping.} \\
\bottomrule
\end{tabular}
\end{table}

\begin{table}[H]
\centering
\caption{ALFWorld, Qwen3-4B-Instruct-2507, training step 200. Both clusters contain the displayed closed-cabinet opening behavior; success cluster $S_1$ filters failure cluster $F_0$.}
\label{tab:alfworld-paired-case}
\setlength{\tabcolsep}{4pt}
\renewcommand{\arraystretch}{1.02}
\begin{tabular}{@{}>{\raggedright\arraybackslash}p{0.14\linewidth}>{\raggedright\arraybackslash}p{\dimexpr0.86\linewidth-8pt\relax}@{}}
\toprule
Cluster & Trajectory behavior and outcome \\
\midrule
\textbf{Success}\par $S_1$ & \begin{tabular}[t]{@{}>{\raggedright\arraybackslash}p{\linewidth}@{}}
\textbf{Behavior 1: finding a cup.} Observe that cabinet 1 is closed; open it to check for the cup. Action: \texttt{open cabinet 1}.  The cabinet is empty. The trajectory succeeds. \\[2pt]
\midrule
\textbf{Behavior 2: finding bread.} Observe that cabinet 1 is closed; open it to check for bread. Action: \texttt{open cabinet 1}.  The cabinet contains a plate. The trajectory succeeds. \\
\end{tabular} \\[3pt]
\midrule
\textbf{Failure}\par $F_0$ & \begin{tabular}[t]{@{}>{\raggedright\arraybackslash}p{\linewidth}@{}}
\textbf{Behavior 1: finding a cup.} Open the closed cabinet to check for the cup. Action: \texttt{open cabinet 1}. \textbf{Later failure:} Cool and place a \texttt{mug} instead of the required \texttt{cup}; then repeatedly look around or check inventory until the 50-turn limit. \\[2pt]
\midrule
\textbf{Behavior 2: finding a pot.} Open the closed cabinet to check for the pot. Action: \texttt{open cabinet 1}. \textbf{Later failure:} Revisit searched locations without checking stoveburner 2, where a successful trajectory finds the pot. Exhaust 50 turns without acquiring or cooling it. \\
\end{tabular} \\[3pt]
\midrule
\multicolumn{2}{@{}p{\linewidth}@{}}{\textbf{Shared pattern.} When a candidate cabinet is closed, open it to inspect its contents for the target object.} \\[3pt]
\multicolumn{2}{@{}p{\linewidth}@{}}{\textbf{Memory decision.} \textbf{$S_1$ filters $F_0$.}  The same closed-cabinet opening behavior occurs in successful trajectories. The corresponding failure cluster is excluded from additional negative shaping.} \\
\bottomrule
\end{tabular}
\end{table}

In AppWorld, the failed paths diverge only after the shared API lookup. The ALFWorld case likewise concerns a shared information-gathering action: opening a closed cabinet to search for a target object.
These cases expose a failure of outcome-only memory: an action can recur in failed trajectories while also supporting successful ones. By vetoing such shared patterns before shaping, PEARL avoids turning terminal failure into an extra penalty on reasonable preparatory behavior.

\section{Conclusion}
\label{sec:conclusion}

We introduced \method{} to turn sparse outcome feedback into selective local supervision for multi-turn agents without training an additional critic.
Rather than assigning extra blame to every decision in a failed trajectory, \method{} stores recurring failure-side patterns, uses successful experience to veto patterns shared across outcomes, and strengthens negative updates only where a retained pattern matches.
Across AppWorld and ALFWorld, experiments at two model scales show improvements over the underlying optimizer, while ablations and bank diagnostics support the roles of targeted shaping, success disambiguation, and recent memory.
These findings suggest that outcome-level feedback becomes more informative when past decisions are selected for reuse, without requiring a locally verified error label for every action.

\clearpage
\appendix
\begingroup
\setlength{\intextsep}{8pt plus 2pt minus 2pt}
\setlength{\textfloatsep}{12pt plus 2pt minus 2pt}
\setlength{\floatsep}{8pt plus 2pt minus 2pt}
\section{Additional Experimental Details}
\label{app:details}

\subsection{Default training configuration}

We train PEARL with Qwen3-4B-Instruct-2507 (4B) and Qwen3-8B (8B) for 300 updates on ALFWorld and 400 updates on AppWorld. Each update uses eight task prompts and 16 trajectories per prompt, for a global batch of 128 trajectories. ARPO starts with eight trajectories per prompt and allocates the remaining rollout budget through branching, with one branch per trigger.

Rollouts allow up to 50 agent turns on ALFWorld and 40 on AppWorld, with a response limit of 4,096 tokens per turn. The context limit for the accumulated interaction is 65,536 tokens for 4B models and 32,768 tokens for 8B models in both environments. Training uses sampling temperature 1.0 and top-$p=1.0$, without top-$k$ truncation.

All four configurations use Adam with a constant learning rate of $10^{-6}$, weight decay 0.1, and $(\beta_1,\beta_2)=(0.9,0.98)$. They use GRPO advantages, symmetric PPO clipping at 0.2, and an entropy coefficient of zero. The branching entropy probe uses 20 tokens, top-$k=128$, $\alpha=0.5$, $\beta=0.2$, and threshold 0.5.

Evaluation uses a 4,096-token response limit and three greedy samples per task (top-$k=1$). The evaluation curves in Figure~\ref{fig:eval-trajectories} include checkpoints at 25-update intervals. ALFWorld uses the Seen and Unseen validation splits, and AppWorld uses the Normal and Challenge test splits. ALFWorld assigns reward 1 for task success and disables the additional invalid-action penalty. The environment prompts and action formats are given in Appendix~\ref{app:prompts}.

The default PEARL banks use HDBSCAN minimum cluster size 10, minimum samples 5, a 25-update window, and a reclustering interval of 10 updates.
Failure shaping is enabled, success shaping is disabled, the shaping temperature is 1.0, and hard success-overlap filtering is used.
The code-level shaping-strength argument is an enable/disable gate in the current implementation; the exponential multiplier itself follows Equation~\ref{eq:shaping}.
All four configurations use counterfactual bank labels derived from unequal-outcome parent--branch pairs. The 0.3 failure/success selection ratios retained in the ALFWorld launchers are legacy settings for the alternative turn-selection path; counterfactual labeling uses the paired regions directly.

\subsection{Process diagnostics}

\paragraph{Unlimited-window bank evolution.}
Table~\ref{tab:w0-process} shows that a growing failure bank does not consistently produce shaping-active matches under unlimited history: retained clusters vary across the snapshots, and retention alone does not guarantee a matching query.

\begin{table}[H]
\centering
\caption{Failure- and success-bank snapshots from the AppWorld hard-filter MCS10/MS5 run with unlimited history ($W=\infty$). ``Retained entries'' counts entries in retained failure clusters, not the number of clusters.}
\label{tab:w0-process}
\setlength{\tabcolsep}{4pt}
\renewcommand{\arraystretch}{1.12}
\begin{tabular*}{\linewidth}{@{\extracolsep{\fill}}rrrrrrr@{}}
\toprule
Step & \shortstack{Failure\\entries} & \shortstack{Success\\entries} & \shortstack{Retained failure\\clusters} & \shortstack{Retained\\entries} & \shortstack{Filtered\\clusters} & \shortstack{Success\\clusters} \\
\midrule
100 & 297  & 347  & 0 & 0    & 2 & 2 \\
200 & 716  & 798  & 1 & 19   & 2 & 2 \\
300 & 1106 & 1195 & 0 & 0    & 5 & 4 \\
400 & 1368 & 1509 & 1 & 1200 & 1 & 15 \\
\bottomrule
\end{tabular*}
\end{table}

\FloatBarrier
\section{Case studies of selective failure memory}
\label{app:pearl-case-studies}

We examine retained failure-side patterns in Qwen3-4B-Instruct-2507 trained with PEARL. The AppWorld and ALFWorld success-veto cases appear in Tables~\ref{tab:appworld-paired-case} and~\ref{tab:alfworld-paired-case} of Section~\ref{sec:memory-case-study}. The immediate environment responses in the retained examples below provide post-hoc audit evidence, not labels available to PEARL when retaining a cluster.

\Needspace{16\baselineskip}
\subsection{Retained patterns in AppWorld and ALFWorld}
Table~\ref{tab:retained-memory-cases} shows two failure-side clusters that remain eligible for shaping: unauthenticated alarm queries in AppWorld and unavailable lamp-inspection commands in ALFWorld. The API and action-validity responses are post-hoc audit observations, not labels used by PEARL to retain either cluster.

\begin{table}[H]
\centering
\caption{Retained failure-side patterns in Qwen3-4B-Instruct-2507 runs. The two rows report separate bank snapshots and two audited decisions per environment.}
\label{tab:retained-memory-cases}
\setlength{\tabcolsep}{4pt}
\renewcommand{\arraystretch}{1.08}
\begin{tabular}{@{}>{\raggedright\arraybackslash}p{0.16\linewidth}>{\raggedright\arraybackslash}p{\dimexpr0.84\linewidth-8pt\relax}@{}}
\toprule
Environment and cluster & Audited decisions, observed responses, and bank decision \\
\midrule
\textbf{AppWorld}\par step 200\par $F_1$ &
\textbf{Delay alarm by one hour:} call \texttt{apis.phone.show\_alarms()} without an access token; the API returns HTTP 401 (missing, invalid, or expired authorization).\par
\textbf{Delay alarm by twenty minutes:} make the same unauthenticated call; the API again returns HTTP 401.\par
\textbf{Matched pattern:} invoke a protected API without satisfying its authentication requirement. \textbf{Bank decision:} retained because the success bank does not veto $F_1$. \\
\midrule
\textbf{ALFWorld}\par step 300\par $F_0$ &
\textbf{After turning on the lamp:} issue \texttt{examine desklamp 1}, which is outside the admissible set; the environment replies ``Invalid action.''\par
\textbf{After a rejected inspection:} following a rejected \texttt{look under desklamp 1}, issue \texttt{examine desklamp 1} and again receive ``Invalid action.''\par
\textbf{Matched pattern:} use an unavailable lamp-inspection command. \textbf{Bank decision:} retained because no success cluster has formed to veto $F_0$. \\
\bottomrule
\end{tabular}
\end{table}

In these sampled traces, immediate rejection feedback helps interpret the local decisions, but PEARL does not use that feedback to assign error labels. Retention only leaves later matching decisions with negative advantage eligible for stronger negative shaping; it does not establish that a pattern caused the final failure.

\section{Agent Prompts for ALFWorld and AppWorld}
\label{app:prompts}

This section records the agent-facing prompt templates used for the two environments. Braced fields are filled at runtime; the text below preserves the wording of the templates. The model-specific chat template supplies the role delimiters around these messages.

\lstdefinestyle{agentprompt}{
  basicstyle=\rmfamily,
  breaklines=true,
  breakatwhitespace=false,
  columns=fullflexible,
  keepspaces=true,
  showstringspaces=false
}
\tcbset{agentpromptbox/.style={
  breakable,
  colback=white,
  colframe=black!55,
  colbacktitle=black,
  coltitle=white,
  fonttitle=\bfseries,
  boxrule=0.4pt,
  arc=8pt,
  outer arc=8pt,
  titlerule=0pt,
  left=10pt,
  right=10pt,
  top=7pt,
  bottom=7pt,
  toptitle=2pt,
  bottomtitle=2pt,
  before skip=8pt,
  after skip=8pt,
  listing only,
  listing engine=listings,
  listing options={style=agentprompt}
}}

\subsection{ALFWorld prompt}
\label{app:alfworld-prompt}

ALFWorld presents one user message per turn. The \texttt{observation} field contains the current textual observation (including the task at the initial turn), and \texttt{admissible\_actions} contains the current environment-provided commands, formatted as a comma-separated list of quoted strings. The agent reasons and returns one command inside \texttt{<action>} tags. The same template is reused after every valid action with updated fields.

\begin{tcblisting}{agentpromptbox,title=ALFWorld user prompt}
You are an expert agent operating in the ALFRED Embodied Environment.
Your current observation is: {observation}
Your admissible actions of the current situation are: [{admissible_actions}].

Now it's your turn to take an action.
You should first reason step-by-step about the current situation, then choose an admissible action for current step and present it within <action> </action> tags.
\end{tcblisting}

If the response lacks a valid tagged action or names a command outside the admissible set, the next user message instead reads:
\begin{tcblisting}{agentpromptbox,title=ALFWorld invalid-action feedback}
Invalid action. Use the <action> </action> format and select one of the admissible actions.
Your admissible actions are: [{admissible_actions}].
\end{tcblisting}

\subsection{AppWorld prompt}
\label{app:appworld-prompt}

AppWorld uses the following full prompt template, including a worked Spotify example and the rules preceding the actual task. The Jinja fields \texttt{main\_user.*}, \texttt{app\_descriptions}, and \texttt{instruction} are filled from the task environment. The \texttt{USER:} and \texttt{ASSISTANT:} markers are parsed into chat messages; the final user message gives the task. After each code execution, the next user message contains the returned output in a fenced block prefixed by \texttt{Output:}.

The Spotify demonstration in the archived template contains a non-executable password lookup: its list comprehension produces Boolean values, then indexes the first Boolean as a dictionary. We reproduce the template as used in the reported runs below. The intended executable lookup would be:
\begin{lstlisting}[basicstyle=\ttfamily,columns=fullflexible,keepspaces=true]
spotify_password = [
    account_password
    for account_password in passwords
    if account_password["account_name"] == "spotify"
][0]["password"]
\end{lstlisting}

\begin{tcblisting}{agentpromptbox,title=AppWorld prompt template,title after break=AppWorld prompt template (continued)}
USER:
I am your supervisor, and you are an AI Assistant whose job is to complete my day-to-day tasks fully autonomously.

To do this, you will need to interact with app(s) (e.g., spotify, venmo etc) using their associated APIs on my behalf. For this you will undertake a *multi-step conversation* using a python REPL environment. That is, you will write the python code, the environment will execute it and show you the result, based on which, you will write python code for the next step and so on, until you've achieved the goal. This environment will let you interact with app(s) using their associated APIs on my behalf.

Here are three key APIs that you need to know to get more information

# To get a list of apps that are available to you.

```python
print(apis.api_docs.show_app_descriptions())
```

# To get the list of APIs under any app listed above, e.g. spotify

```python
print(apis.api_docs.show_api_descriptions(app_name='spotify'))
```

# To get the specification of a particular api, e.g. spotify app's login api

```python
print(apis.api_docs.show_api_doc(app_name='spotify', api_name='login'))
```

Each code execution will produce an output that you can use in subsequent calls. Using these APIs, you can now generate code, that I will execute, to solve the task. Let's start with the task

My name is: {{ main_user.first_name }} {{ main_user.last_name }}. My personal email is {{ main_user.email }} and phone number is {{ main_user.phone_number }}.
Task: How many playlists do I have in my Spotify playlist library?

ASSISTANT:
Okay. Let's first find which APIs are available to use in Spotify.

```python
print(apis.api_docs.show_api_descriptions(app_name='spotify'))
```

USER:
Output:
```
[
 ...
 "login : Login to your account.",
 "logout : Logout from your account.",
 ...
]
```

ASSISTANT:
Okay. Looks like I can use the `login` api. Let's find its specifications.

```python
print(apis.api_docs.show_api_doc(app_name='spotify', api_name='login'))
```

USER:
Output:
```
{
 "app_name": "spotify",
 "api_name": "login",
 "path": "/spotify/auth/token",
 "method": "POST",
 "description": "Login to your account.",
 "parameters": [
  {
   "name": "username",
   "type": "string",
   "required": true,
   "description": "Your account email.",
   "default": null,
   "constraints": []
  },
  {
   "name": "password",
   "type": "string",
   "required": true,
   "description": "Your account password.",
   "default": null,
   "constraints": []
  }
 ],
 "response_schemas": {
  "success": {
   "access_token": "string",
   "token_type": "string"
  },
  "failure": {
   "message": "string"
  }
 }
}
```

ASSISTANT:
I need the supervisor's username and password. Let's see if any app can help me get that.

```python
print(apis.api_docs.show_app_descriptions())
```

USER:
Output:
```
{{ app_descriptions }}
```

ASSISTANT:
Looks like the supervisor app could help me with that.

```python
print(apis.api_docs.show_api_descriptions(app_name='supervisor'))
```

USER:
Output:
```
[
 ...
 "show_account_passwords : Show your supervisor's account passwords."
 ...
]
```

ASSISTANT:
I can use `show_account_passwords` to get the passwords. So I will call that and save it in a variable for future reference.

```python
passwords=apis.supervisor.show_account_passwords()
print(passwords)
```

USER:
Output:
```
[
 {
  "account_name": "spotify",
  "password": "dummy_spotify_pass"
 },
 {
  "account_name": "file_system",
  "password": "dummy_fs_pass"
 },
 ...
]
```

ASSISTANT:
So the Spotify password is an entry in the `passwords` list with the account_name=spotify. I can use it to login now with the provided email address: {{ main_user.email }}.

```python
spotify_password = [account_password["account_name"] == "spotify" for account_password in passwords][0]["password"]
login_result = apis.spotify.login(username='{{ main_user.email }}', password=spotify_password)
print(login_result)
```

USER:
Output:
```
{
 "access_token": "fake_access_token",
 "token_type": "Bearer"
}
```

ASSISTANT:
First, let's check the available APIs for the Spotify app.

```python
print(apis.api_docs.show_api_descriptions(app_name='spotify'))
```

USER:
Output:
```
[
 ...
 {
  "name": "show_playlist_library",
  "description": "Search or show a list of playlists in your playlist library."
 },
 {
  "name": "show_playlist",
  "description": "Get detailed information about a specific playlist. You can view your own playlists or others' playlists if they are public."
 },
 ...
]
```

ASSISTANT:
The `show_playlist_library` API seems to be the one we need. Let's check its specifications.

```python
print(apis.api_docs.show_api_doc(app_name='spotify', api_name='show_playlist_library'))
```

USER:
Output:
```
{
 "app_name": "spotify",
 "api_name": "show_playlist_library",
 "path": "/spotify/library/playlists",
 "method": "GET",
 "description": "Search or show a list of playlists in your playlist library.",
 "parameters": [
  {
   "name": "access_token",
   "type": "string",
   "required": true,
   "description": "Access token obtained from spotify app login.",
   "default": null,
   "constraints": []
  },
  {
   "name": "query",
   "type": "string",
   "required": false,
   "description": "The search query string.",
   "default": "",
   "constraints": []
  },
  {
   "name": "is_public",
   "type": "boolean",
   "required": false,
   "description": "Whether to show public playlists or private playlists.",
   "default": null,
   "constraints": []
  },
  {
   "name": "page_index",
   "type": "integer",
   "required": false,
   "description": "The index of the page to return.",
   "default": 0,
   "constraints": [
    "value >= 0"
   ]
  },
  {
   "name": "page_limit",
   "type": "integer",
   "required": false,
   "description": "The maximum number of results to return per page.",
   "default": 5,
   "constraints": [
    "value >= 1, <= 20"
   ]
  },
  {
   "name": "sort_by",
   "type": "string",
   "required": false,
   "description": "The attribute to sort the library playlists by prefixed with +/- to reflect ascending/descending. Valid attributes: created_at and title. ...",
   "default": null,
   "constraints": []
  }
 ],
 "response_schemas": {
  "success": [
   {
    "playlist_id": 1,
    "title": "string",
    "is_public": true,
    "rating": 0.0,
    "like_count": 1,
    "review_count": 1,
    "owner": {
     "name": "string",
     "email": "user@example.com"
    },
    "created_at": "2019-01-01T00:00:00",
    "song_ids": [
     1
    ]
   }
  ],
  "failure": {
   "message": "string"
  }
 }
}
```

ASSISTANT:
I need to page through all the playlists to get the list of playlists and save it in `playlists`.

```python
page_index = 0
playlists = []
while page_index < 10:
  playlist_page = apis.spotify.show_playlist_library(access_token=spotify_access_token, page_index=page_index)
  if playlist_page:
    playlists.extend(playlist_page)
    page_index += 1
  else:
    break
num_playlists = len(playlists)
print(num_playlists)
```

USER:
Output:
```
23
```

ASSISTANT:
Now that the task is completed, I can call apis.supervisor.complete_task(). Since this task has an answer to be returned, I will pass that as an argument.

```python
apis.supervisor.complete_task(answer=num_playlists)
```

USER:
Output:
Marked the active task complete.

USER:
**Key instructions**:

A. General instructions:

- Act fully on your own. You must make all decisions yourself and never ask me or anyone else to confirm or clarify. Your role is to solve the task, not to bounce questions back, or provide me directions to follow.
- You have full access -- complete permission to operate across my connected accounts and services.
- Never invent or guess values. For example, if I ask you to play a song, do not assume the ID is 123. Instead, look it up properly through the right API.
- Never leave placeholders; don't output things like "your_username". Always fill in the real value by retrieving it via APIs (e.g., Supervisor app for credentials).
- When I omit details, choose any valid value. For example, if I ask you to buy something but don't specify which payment card to use, you may pick any one of my available cards.
- Avoid collateral damage. Only perform what I explicitly ask for. Example: if I ask you to buy something, do not delete emails, return the order, or perform unrelated account operations.

B. App-specific instructions:

- All my personal information (biographical details, credentials, addresses, cards) is stored in the Supervisor app, accessible via its APIs.
- Any reference to my friends, family or any other person or relation refers to the people in my phone's contacts list.
- Always obtain the current date or time, from Python function calls like `datetime.now()`, or from the phone app's get_current_date_and_time API, never from your internal clock.
- All requests are concerning a single, default (no) time zone.
- For temporal requests, use proper time boundaries, e.g., when asked about periods like "yesterday", use complete ranges: 00:00:00 to 23:59:59.
- References to "file system" mean the file system app, not the machine's OS. Do not use OS modules or functions.
- Paginated APIs: Always process all results, looping through the page_index. Don't stop at the first page.

C. Code-operation instructions

- Make sure to end code blocks with ``` followed by a newline(\n).
- Remember, you can use the variables in your code in subsequent code blocks.
- Remember that the email addresses, access tokens and variables (e.g. spotify_password) in the example above are not valid anymore.
- Always look at API specifications (using apis.api_docs.show_api_doc) before calling an API.
- Write small chunks of code and only one chunk of code in every step. Make sure everything is working correctly before making any irreversible changes.
- The Python environment supports the standard library. But system-level operations that may access or affect OS files, processes, etc., are not allowed and will raise an error if called.
- To interact with apps, only use the provided app APIs, and not the corresponding Python packages, e.g., do NOT use `spotipy` for Spotify.
- The provided API documentation has both the input arguments and the output JSON format. Use this information when making API calls and parsing their outputs.

D. Task-completion instructions:

You must call the `apis.supervisor.complete_task` API after completing the task.
- If an answer is needed, e.g., for "How many songs are in the Spotify queue?", call it with the appropriate answer argument value.
- If no answer is required, e.g., for "Start my Spotify music player.", omit the answer argument (or set it to None/null).
- The task is doable, but if you cannot find a way, you can call it with status="fail" to exit with failure.

When the answer is given:
- Keep answers minimal. Return only the entity, number, or direct value requested - not full sentences.
  E.g., for the song title of the current playing track, return just the title.
- Numbers must be numeric and not in words.
  E.g., for the number of songs in the queue, return "10", not "ten".

USER:
Using these APIs, now generate code to solve the actual task:

My name is: {{ main_user.first_name }} {{ main_user.last_name }}. My personal email is {{ main_user.email }} and phone number is {{ main_user.phone_number }}.
Task: {{ instruction }}
\end{tcblisting}

\endgroup

\end{document}